\RequirePackage{fix-cm}
\documentclass[twocolumn]{svjour3}

\smartqed  

\usepackage{graphicx}
\usepackage{amsmath}
\usepackage{amsfonts}
\usepackage{amssymb}
\usepackage{subfig}
\usepackage{adjustbox}
\usepackage{ragged2e}
\usepackage{hhline}
\usepackage{multicol}
\usepackage{float}
\usepackage{multirow}
\usepackage{mathptmx}      
\usepackage{algorithm}
\usepackage{algpseudocode}
\usepackage{latexsym}

\usepackage[colorlinks=true,breaklinks=true,bookmarks=true,urlcolor=blue,citecolor=blue,linkcolor=blue,bookmarksopen=false,draft=false]{hyperref}

\usepackage[sort,compress,numbers]{natbib}

\usepackage{flushend}
\graphicspath{{./}}

\begin{document}

\title{SE-MD: A Single-encoder multiple-decoder deep network for point cloud generation from 2D images}



\author{Abdul Mueed Hafiz \and Rouf Ul Alam Bhat \and Shabir Ahmad Parah	\and M. Hassaballah}


\institute{Abdul Mueed Hafiz  \at
              Department of Electronics \& Communication Engineering,\\ Institute of Technology, University of Kashmir,\\ Srinagar, J\&K, 190006, India\\
              Tel.: +91-7006474254\\
              \email{mueedhafiz@uok.edu.in}           
           \and
	Rouf Ul Alam Bhat \at
              Department of Electronics \& Communication Engineering,\\ Institute of Technology, University of Kashmir,\\ Srinagar, J\&K, 190006, India\\
           \and
	 Shabir Ahmad Parah \at
            Department of Electronics and Instrumentation Technology,\\ University of Kashmir,  Srinagar, J\&K, 190006, India\\
	\and
	 M. Hassaballah \at
           Department of Computer Science,\\ Faculty of Computers and Information,\\ South Valley University, Qena, 83523, Egypt\\
}


\maketitle

\begin{abstract}
3D model generation from single 2D RGB images is a challenging and actively  researched computer vision task. Various techniques using conventional network architectures have been proposed for the same. However, the body of research work is limited and there are various issues like using inefficient 3D representation formats, weak 3D model generation backbones, inability to generate dense point clouds, dependence of post-processing for generation of dense point clouds, and dependence on silhouettes in RGB images. In this paper, a novel 2D RGB image to point cloud conversion technique is proposed, which improves the state of art in the field due to its efficient, robust and simple model by using the concept of parallelization in network architecture. It not only uses the efficient and rich 3D representation of point clouds, but also uses a novel and robust point cloud generation backbone in order to address the prevalent issues. This involves using a single-encoder multiple-decoder deep network architecture wherein each decoder generates certain fixed viewpoints. This is followed by fusing all the viewpoints to generate a dense point cloud. Various experiments are conducted on the technique and its performance is compared with those of other state of the art techniques and impressive gains in performance are demonstrated. Code is available at \url{https://github.com/mueedhafiz1982/}


\end{abstract}

\section{Introduction}
Three-dimensional model generation from a single RGB image \cite{wong20103d,chowdhury2021fixed,rc_sur1,rc1} has been around for some time and is quite a challenge in computer vision \cite{hassaballah2019recent}. Its importance lies in the fact that this task represents one of the fundamental goals of computer vision (i.e., interpretation or understanding of scenic vision) \cite{kim2021method}. Human vision system is an expert in the interpretation of stereo vison based images and the above mentioned task represents this aspect of the former \cite{rc_sur2,khan2020learning}. In spite of the fact that some progress has been made recently using deep network based models and some large-scale databases are available for research in this field, generation of fine 3D geometry for numerous categories of objects with different topologies is still challenging \cite{fu2021single}. One popular 3D representation for tackling this challenge is the point cloud due to a number of important advantages \cite{wang2014construction,texture}. For example, it conveniently represents varying geometry unlike a mesh. It does not have problems with cubic complexity unlike a 3D voxel-grid. Also, it can be used for shape reconstruction of implicit functions with the help of single evaluations in neural networks \cite{rc0,rc05}.

In spite of their success, 3D convolutional networks have the intrinsic drawback when they model a volumetric shape representation \cite{original}. A 2D image has pixels with rich spatial and texture information which is not the case with a volumetric representation which has sparse information. Also when a voxel-grid is used to express a 3D object, a voxel which is outside or inside the object is not important and is not much useful. Alternately, the most useful data for 3D object representation is the surface-based data which is present scantily in the voxel occupancy grid. As a result, 3D convolutional networks are very wasteful computationally and memory-wise for prediction of data, because they use very complex 3D convolution mechanisms, which in turn have severe limitations on 3D volumetric shape granularity generated using high-end GPUs. 

There are various existing techniques for 2D image to 3D model conversion having different strategies and different backbone networks. However, many of these techniques rely on inefficient and wasteful representative formats like meshes \cite{t27,t16} and voxel-grids \cite{o2}. Some techniques have weak backbones which generate sparse point clouds \cite{pyramid}. This issue in turn necessiates complex and computationally expensive post-processing for generation of dense point clouds. Some techniques rely on silhouttes of objects found in RGB images \cite{silhoutte} which is not how expert vision systems like that of humans work, the realization of the latter being the goal of this computer vision task. Some techniques use the ground-truth (GT) point clouds of the training data as co-input in addition to RGB images \cite{lmnet}. Again this procedure is unfound in the human vision system. In our opinion, staying on the main path is the best approach and any detours may be unsuccessful. The need of the hour is a 3D model generation system which addresses these issues. Accordingly, a 2D RGB image to point cloud conversion technique is proposed in this paper which uses point clouds for 3D representation, takes only 2D RGB images as input, has an efficient parallelized backbone network with slight computational overhead in comparison to single stream networks, does not have a post-processing pipeline,... etc. Further, this architecture can be extended to other computer vision tasks which can benefit from its advantages \cite{hassaballah2020deep}.

In this paper, a novel technique is proposed for 3D point cloud generation from single RGB images using a single-encoder multiple-decoder deep network. It is based on the parallelization concept \cite{paa_ens} introduced into the backbone network of Lin et al. \cite{original}, which leads to impressive gains in performance. The proposed technique is capable of being trained on multiple categories of objects. The main focus of the proposed technique is to address the problem of 2D RGB image to point cloud conversion by introducing a new deep network architecture, which is efficient, robust, convenient to train, and is relatively convenient to implement. It has two stages; in the first stage, $N$ viewpoint images are obtained from a single RGB image using the proposed network, while in the next stage, these images are fused to reconstruct the point cloud. Since the proposed technique uses 2D convolution, it conveniently achieves high resolution point clouds without having the issues of 3D convolution. Also, the proposed deep network does not use point clouds as co-input hence it is a 'pure' 2D image to point cloud conversion technique as is the human vision system which does not use point clouds and depends only on stereo images for visual environment understanding. Hence, it goes in the natural direction and does not suffer from additional complexity. Evaluation of the proposed technique has been done in different ways. Both single-category as well as multiple-category data have been used for training the model and testing. It is empirically demonstrated that the proposed technique generalizes better than other competing techniques. Additionally, the proposed technique does not need 2.5D or 3D data and does not have the issues related to processing such data. To summarize, the main contributions of the proposed technique are given below:

\begin{itemize}
	\item A new technique for generation of a 3D shape in the form of a point cloud from a single 2D RGB image using a two-stage technique with first generation of  intermediate fixed coordinate images and second reprojection by fusion to form the point cloud based 3D representation of the object.
	\item To the best of our knowledge, it is the first work to use a parallelization in point cloud generation backbones featuring single-encoder and multiple-decoder deep network for a computer vision task and demonstrates the efficacy of the model, which can be extended to other computer vision tasks.
	\item Our technique reconstructs 3D representations better than previous techniques on datasets like ShapeNet \cite{shapenet} and impressive gains in performance are witnessed.
\end{itemize}

The rest of the paper is organized as follows. Section \ref{Relatedwork} discusses the related work regarding the focus of this paper (i.e., 3D model generation using single 2D image). Section \ref{proposed} provides details of the proposed technique for generating 3D representations with fine grained point clouds. Section \ref{Exp}
presents the experimental results and evaluation analysis of the proposed technique. 
Finally, Section \ref{Con} concludes this paper.

\section{Related work}
\label{Relatedwork}

With the development of deep learning models, single 2D image to 3D model generation has become an active area of research with substantial progress. Due to regularity, initial works chiefly used to learn the reconstruction of voxel grids using a 3D supervised technique \cite{o2} or a 2D supervised technique \cite{o24} with the help of a differentiable renderer \cite{t29,t25}. In spite of this, these techniques are able to reconstruct low resolution shapes e.g. 32 or 64, because of the complex cubic nature of the voxel grids. Additionally, many techniques \cite{t8,t24} have been developed for increasing the model resolution, however these techniques are too complicated to be followed. Mesh-based techniques \cite{t27,t16}  are an alternative for increasing the 3D model resolution. In spite of this, these techniques face difficulty in handling of arbitrary topologies because the vertex-based topology for the generated 3D shapes chiefly inherits from the template. Point cloud representation techniques \cite{t7,t19,t32,lmnet} provide a viable means for single 2D RGB image to 3D model conversion. There have been issues with resolution. However the proposed technique is able to address these issues. 

There are also techniques that map the texture of the object from the 2D image to the 3D representation using meshes \cite{t13} or point clouds \cite{texture,t13,t21,t34,o24,t33,t32}. However, texture based 3D model generation research is still in early phase and has been applied to artificial images and is not capable of texture generation on real images e.g. like those found in databases like Pix3D \cite{pix3d} . They also have issues like complex pipelines, difficult training, need for large computational resources. Hence, instead of focusing on improving the crucial area of 3D model generation they currently tend to take the focus away from it. 

Shape completion is another upcoming area in image to 3D model generation e.g. in view of occlusion \cite{paa_occ}. It involves inferring the complete 3D geometrical model from partial observations. Various techniques use voxel grids \cite{t5} or point clouds \cite{t31,t30,t1} for shape completion with backbones like the PointNet architecture \cite{t18}. Although these techniques demonstrate decent shape completion, they have limited resolution. The body of work is small and the shape completion task can be considered as a post-processing in 2D image to 3D model conversion tasks which are not efficient enough. 

We focus on the backbone of the above computer vision task for making the main task efficient, robust and simple by introducing parallelization in decoder portion. Of course, post processing techniques may be added to the 3D models generated by the proposed technique. One more important aspect of the proposed technique is that it falls within the category of ‘pure' 2D image based 3D model generation techniques as found in works of Lin et al. \cite{original}, which do not depend on point cloud data as co-input. The human vision process also depends only on stereo-vision based captured images and is very efficient for real-time depth estimation. It should be noted that this novel architecture which is a first in a computer vision task, can be extended in other tasks.

\section{The proposed technique}
\label{proposed}

\begin{figure*}
	\centering
\includegraphics[width=14 cm]{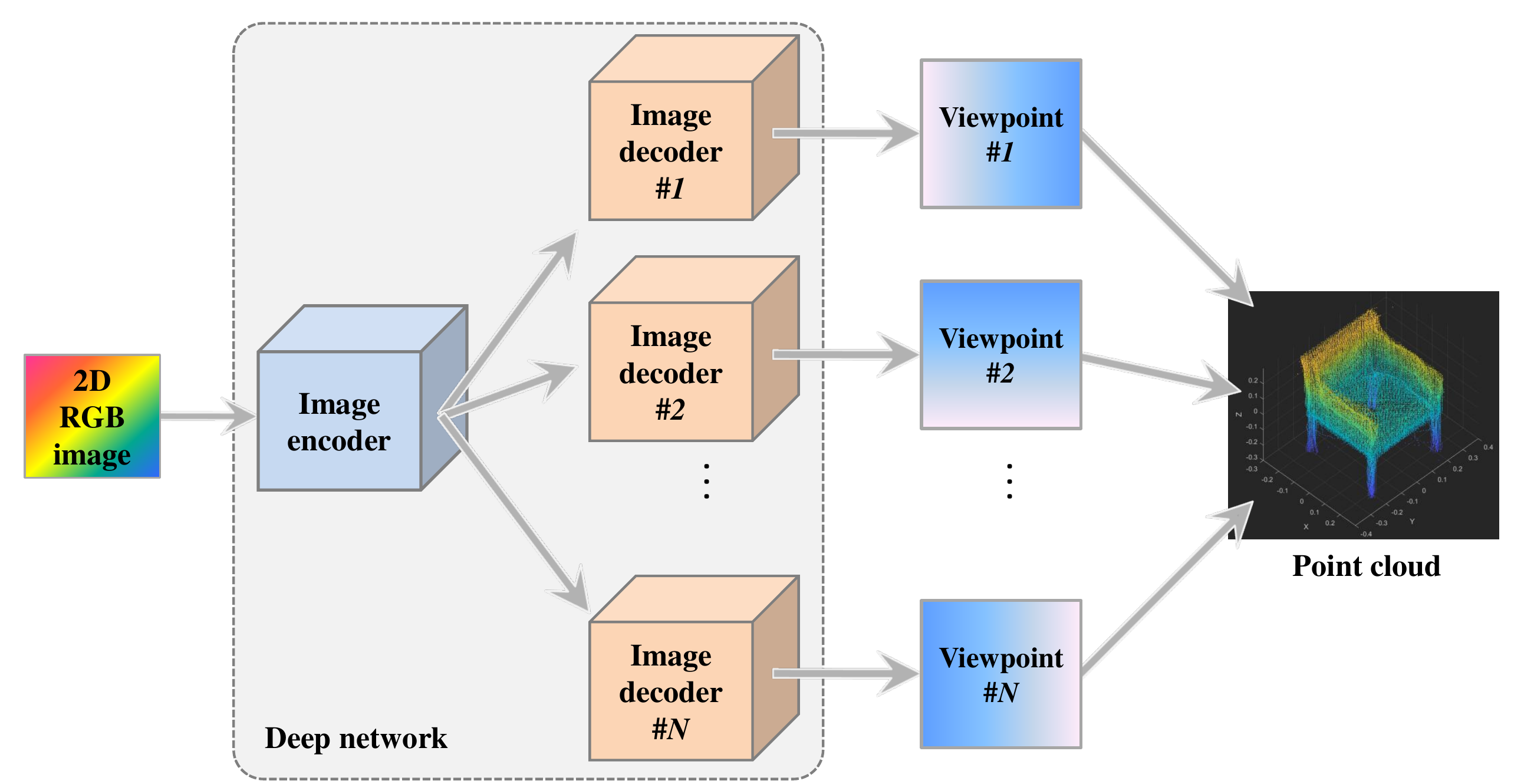}
\caption{General network architecture of the proposed technique.}
\label{fig1}
\end{figure*}

The aim of the proposed technique is generating 3D representations with fine grained point clouds. Inspired by the model used in Lin et al. \cite{original}, which consists of a conventional single-encoder single-decoder deep network, we propose a single-encoder multiple-decoder model as shown in Fig. \ref{fig1}. The 2D image is fed to the encoder that maps the former to a rich representative space. From the representative space, rich point clouds are generated using \textbf{\textit{N}} structure generators based 2D convolution with a criterion for 2D projection. It should be noted that our technique is different than that in Lin et al. \cite{original} because the latter uses a conventional architecture (single-encoder single-decoder network), while the proposed technique for the first time uses a single-encoder multiple-decoder network. Also, the computational overhead introduced due to parallelization is small.


\subsection{\textit{N} Structure generators}
Each structure generator gives a 3D structure prediction of an object present in a 2D RGB image for a single viewpoint, i.e. 3D coordinates as in  \( \hat{x}_{i}= \left[ \hat{x}_{i}~\hat{y}_{i}~\hat{z}_{i} \right] ~^{T} \)  for every pixel location. Values of pixels in 2D images may be created by generative architectures using convolution chiefly because of major dependencies in spatial domains. Such phenomena may be also shown by point clouds if they are treated as multi channel images in a 2D grid having coordinates in the form of \textit{(x, y, z)}. Due to this fact, each structure generator uses 2D convolution for prediction of images having \textit{(x, y, z)} form for representation of 3D surfaces. The above approach prevents issues like heavy time consumption and need for heavy computation hardware for 3D convolution for predicting volumes. 

Considering 3D matrices for transformation for the \textbf{\textit{N }}viewpoints given by \textbf{(R\textsubscript{1}} , \textbf{t\textsubscript{1}) }$ \ldots $  \textbf{(R\textit{\textsubscript{N}}} , \textbf{t\textit{\textsubscript{N}})}, every 3D point represented by  \( \hat{x}_{i} \)  at the \textit{n\textsuperscript{th}} viewpoint can be used for transformation to its 3D coordinate representation  \( \widehat{p_{i}} \)  using  
\begin{equation}
\widehat{p_{i}}=R_{n}^{-1} \left( K^{-1}\hat{x}_{i}-t_{n} \right)   \forall i
\end{equation}
Where \textbf{K} denotes a camera predefined matrix. This is the relationship definition among the generated 3D points and the fusion of all point clouds inside the 3D coordinates and it is the result of the proposed architecture.

\subsection{Optimization of the 2D joint projection}

Pseudo-rendering defined depth images  \( \hat{Z}=f_{PR} \left(  \{ \hat{x}_{i}^{'} \}  \right)  \) are used alongside masks  \( \hat{M} \)  for different viewpoints in order to optimize the system. Loss is defined as a combination of mask loss  \( L_{mask} \)  and depth loss  \( L_{depth} \) ,\ with their respective definitions as  
\begin{equation}
L_{mask}= \sum _{k=1}^{K}-M_{k}\log \hat{M}_{k}- \left( 1-M_{k} \right) \log \hat{M}_{k} 
\end{equation}
\begin{equation}
L_{depth}= \sum _{k=1}^{K} \Vert \hat{Z}_{k}-Z_{k} \Vert _{1}
\end{equation}

Optimization is simultaneously done over different viewpoints. For the \textit{k}\textsuperscript{th} viewpoint, let \( M_{k} \)  and  \( Z_{k} \)  represent the ground truth (GT) image and the depth image respectively. Loss  \( L_{1} \)  is calculated for every depth element. Cross-entropy loss is used for the mask. The combined/holistic loss with  the weight factor \(  \lambda  \) is represented as
\begin{equation}
L=L_{mask}+ \lambda  \cdot L_{depth}
\end{equation}

Optimizing each structure generator over its own \textit{n}\textsuperscript{th} projection followed by combining all of the projections leads to enforcement of combined 3D reasoning for geometry among the generated point clouds obtained from the N viewpoints. The same optimization also ensures uniform distribution of the error among different viewpoints as against concentrating on predefined \textit{N} viewpoints. Algorithm \ref{alg} summarizes main steps of the proposed technique to generate 3D object point clouds.

\begin{algorithm}{
 \caption{A procedure of generating 3D object point clouds using the proposed technique.}
 \label{alg}}
\begin{algorithmic}[1]
 \State \textbf{Input:}   \textit{M} RGB 2D images; \textit{K} = Number of viewpoints; \textit{N} = Number of decoders
 \State \textbf{Output:}3D object point clouds
  \While{$i$ in range: (1, \textit{M})}
  \State	Feed image \textit{i} to single encoder
  \State	Extract feature map \textit {\( f_{i} \)} from encoder
     \For{$j$ in range: (1, \textit{N})}
	  \State Feed feature map \textit {\( f_{i} \)} to decoder \textit{j}	
	  \State Extract \textit{K/N} viewpoints from decoder \textit{j}
	\EndFor  
  \State    Fuse \textit{K} viewpoints to form point cloud for image \textit{i}
\EndWhile
\end{algorithmic}
\end{algorithm}

\subsection{Network architecture}
The encoder and the decoder architectures are similar to those used in Lin et al. \cite{original}. However, \textit{N} identical decoders are used in parallel each feeding on the common feature map generated by the encoder. Various values of \textit{N} (number of decoders) are tried, which are factors of 8 (e.g., 2,4,8) for generation of 4,2,1, viewpoints individually per decoder respectively. A maximum of 8 viewpoints are generated. The structure generator has a single encoder and multiple decoders, each using 2D convolutional layers with 3 \(  \times  \) 3 kernels.  All feature map dimensions are halved during each convolutional encoding and are doubled during each convolutional decoding. Details and dimensions of the encoder and decoder are listed in Table \ref{tab1}. Batch normalization layers \cite{o10} and ReLU layers are added between different layers of the network. Each decoder generates \textbf{8/\textit{N}} images of dimensions 128 \(  \times  \) 128 \(  \times  \) 4 (\textit{x, y, z,} binary mask), where the unique viewpoints are those belong to 8 corners of a centered cube.

\begin{table*}
	\centering
	\caption{Details of the encoder and decoder network architectures.}
	\label{tab1}
	\begin{tabular}{p{0.25in}p{0.4in}p{0.4in}p{1.4in}p{1.4in}}
		\hline
		\multicolumn{1}{|p{0.25in}}{\multirow{1}{*}\Centering {\textbf{Sec.}}} & 
		\multicolumn{1}{|p{0.6in}}{\multirow{1}{*}\Centering {\textbf{Input size}}} & 
		\multicolumn{1}{|p{0.7in}}{\multirow{1}{*}\Centering {\textbf{Latent vector}}} & 
		\multicolumn{2}{|p{2.8in}|}{\Centering \textbf{Number of filters}} \\
		\hhline{~~~--}
		\multicolumn{1}{|p{0.25in}}{} & 
		\multicolumn{1}{|p{0.4in}}{} & 
		\multicolumn{1}{|p{0.4in}}{} & 
		\multicolumn{1}{|p{1.4in}}{\Centering Encoder } & 
		\multicolumn{1}{|p{1.5in}|}{\Centering Decoder} \\
		\hhline{-----}
		\multicolumn{1}{|p{0.25in}}{\Centering 4.2} & 
		\multicolumn{1}{|p{0.4in}}{\Centering 64x64} & 
		\multicolumn{1}{|p{0.4in}}{\Centering 512-D} & 
		\multicolumn{1}{|p{1.4in}}{Conv:96, 128, 192, 256 \par Linear:2048, 1024, 512} & 
		\multicolumn{1}{|p{1.4in}|}{Linear: 1024, 2048, 4096 \par Deconv: 192, 128, 96, 64, 48} \\
		\hhline{-----}
		\multicolumn{1}{|p{0.25in}}{\Centering 4.3} & 
		\multicolumn{1}{|p{0.4in}}{\Centering 128x128} & 
		\multicolumn{1}{|p{0.4in}}{\Centering 1024-D} & 
		\multicolumn{1}{|p{1.4in}}{Conv: 128, 192, 256, 384, 512 \par Linear: 4096, 2048, 1024} & 
		\multicolumn{1}{|p{1.5in}|}{Linear: 2048, 4096, 12800 \par Deconv: 384, 256, 192, 128, 96} \\
		\hhline{-----}
		
	\end{tabular}
\end{table*}

\section{Experimental Results}
\label{Exp}

The proposed technique is evaluated by studying its performance for single 2D image to 3D point cloud conversion followed by comparing it to state of the art techniques.

\subsection{Database}

The ShapeNet database \cite{shapenet} is mainly used because the latter has been used in many point cloud based techniques, and as explained earlier, point clouds have many advantages over other representation models like meshes and voxel-grids. Databases like Pix3D \cite{pix3d} have been scantily used and benchmarking is limited and as such it offers limited scope for pixel cloud based research. Also, techniques using Pix3D such 3D-LMNet and others (See\cite{lmnet,pyramid}) sample limited number pixels from the surface of the ground truth (GT) models for subsequent point cloud representation and experimentation. Although some works like those given in \cite{silhoutte} use Pix3D because of it contains real-world images, however they simultaneously depend on its image silhouettes (additional information) for 3D model generation, which is not an appropriate approach in our opinion because this is different from the natural process of 2D image to 3D model conversion which does not use silhouetting. 

\begin{figure*}[!h]
	\centering
	\includegraphics[width=15 cm]{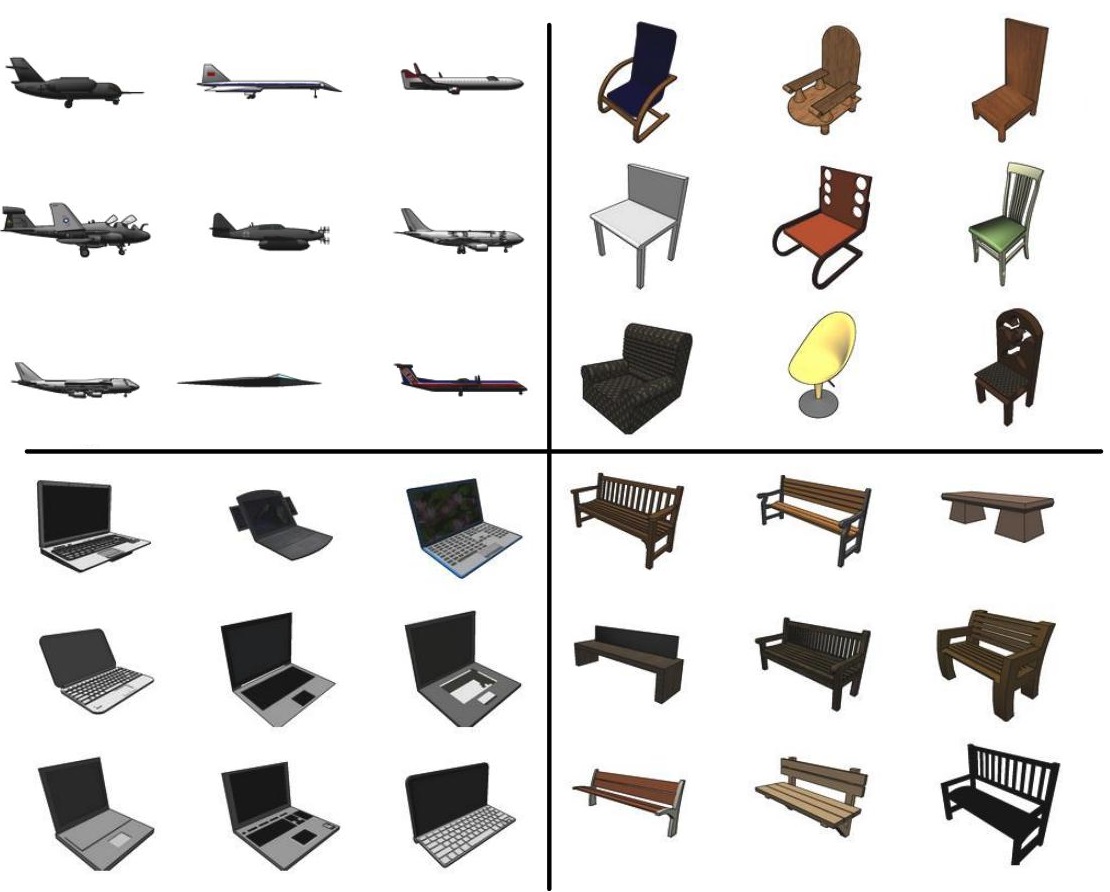}
	\caption{Samples of four aligned models from the ShapeNet database.}
	\label{fig0}
\end{figure*}

The popular ShapeNet database \cite{shapenet} has been used for training and evaluation of all models. It consists of a large number of 3D model categories (about 3 million shapes from online 3D model repositories), where semantic annotations for each 3D model is provided including geometric attributes, consistent rigid alignments, parts and bilateral symmetry planes, and physical sizes. For each model, we use random pre-rendering on 100 depth as well as mask images in pairs for a size of 128 \(  \times  \) 128 for different viewpoints. The inputs to the proposed technique are object instances generated by pre-rendering with fixed elevation and with 24 unique azimuth angles. Fig. \ref{fig0} shows sample images for four 3D aligned models found in the ShapeNet database.

\subsection{Network training}
Adam optimizer \cite{o13} is used for network training. Training has two stages viz. pretraining of the structure generator for prediction of \textit{N} viewpoint depth images. A constant learning rate of 5e-3 is used for the proposed technique respectively for obtaining optimum results. A smaller constant learning rate is used as it was found empirically that multi-stream networks learn better using smaller learning rates as compared to their single stream counterparts. Subsequently, the full network is optimized by fine-tuning it on joint 2D projection. A constant learning rate of 5e-6 is used for the single-stream model. Also \(  \lambda  \) = 1.0 and \textit{K}= 5 are used.

\subsection{Evaluation metrics}
The following metrics have been used for performance evaluation of the proposed technique.

\noindent \textbf{3D Euclidean distance} is measured for every point  \( \widehat{p_{i}} \)  between source 3D model and target 3D model with the distance \textbf{\textit{S}} to the target 3D model, where
\begin{equation}
\varepsilon _{i}=\mathop{\min }_{\mathop{p}_{j}} \Vert \hat{p}_{i}-\hat{p}_{j} \Vert _{2}
\end{equation}

The 3D Euclidean distance metric is bidirectional and it is important to report both its forward and backward values as they represent different quality aspects viz. the forward value gives similarity of 3D shapes and the backward value gives surface coverage. The GT 3D models are represented by 100K densified point clouds.  

\noindent \textbf{Chamfer distance(CD)} between two point-sets  \( \hat{X}_{P} \)  and  \( X_{P} \) is a loss function, which is the nearest-neighbor distance metric and is given by
\begin{equation}
d_{Chamfer} \left( \hat{X}_{P},X_{P} \right) = \sum _{x \in \hat{X}_{P}}^{}\mathop{\min }_{y \in \mathop{X}_{P}} \Vert x-y \Vert _{2}^{2} +   \sum _{y \in \hat{X}_{P}}^{}\mathop{\min }_{x \in \mathop{X}_{P}} \Vert x-y \Vert _{2}^{2}
\end{equation}

\noindent \textbf{Earth mover distance (EMD)} between two point-sets  \( \hat{X}_{P} \)  and  \( X_{P} \)  is given by
\begin{equation}
d_{EMD} \left( \hat{X}_{P},X_{P} \right) =\mathop{\min }_{ \phi :\mathop{\hat{\mathop{X}}}_{P}  \rightarrow  ~\mathop{X}_{P}} \sum _{x \in \hat{X}_{P}}^{} \Vert x- \phi  \left( x \right)  \Vert _{2}
\end{equation}
where  \(  \phi  :\hat{X}_{P}  \rightarrow  ~X_{P} \)  is the bijection. This definition enforces point-wise mapping among two sets and hence ensures uniform point predictions.
\subsection{Performance analysis}
\subsubsection{Single object category}
\begin{figure*}
	\centering
\includegraphics[width=12 cm]{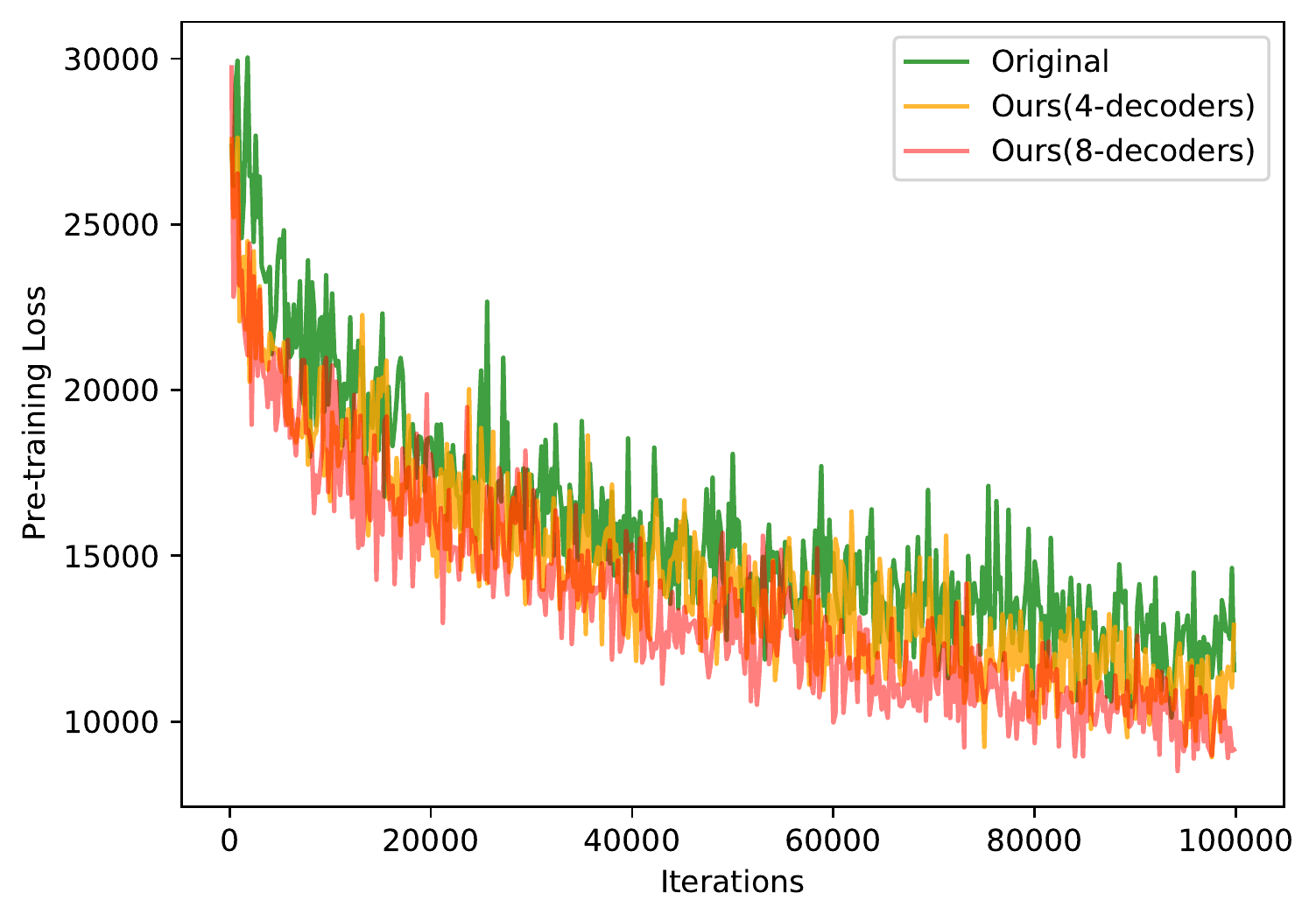}
\caption{Pretraining comparison for the network used in \cite{original} ('Original') and variants of our network (Best viewed in color).}
\label{fig2}
\end{figure*}

The evaluation begins with that of point clouds on 3D generation for single object category viz. chair category of ShapeNet, consisting of 6,778 models. The proposed network is pretrained for 200K iterations and fine-tuned end-to-end for 100K iterations and 80$\%$ -20$\%$ training/testing split is considered. The 3D Euclidean distances (both forward and backward) for the single category experiment were measured for the proposed deep model with single-encoder, and 2-, 4- and 8- decoders. The proposed network variants with 2, 4 and 8 decoders generate 4, 2 and 1 viewpoints per decoder respectively. The related results are shown in Table \ref{tab1_5}. It is clear that as the parallelization increases (i.e. the number of decoders used in the proposed network increases), the performance of the latter increases. For the proposed single-encoder 8-decoder network, both the metrics are better than that of \cite{original}. Fig. \ref{fig2} shows pre-training error plots for \cite{original} and the proposed technique for first 100K iterations. As can be observed from the figure, our pre-training is much more efficient, and also with more parallelization (using 8 decoders instead of 4 decoders in the proposed approach) the pretraining efficiency increases further.


\begin{table*}
	\centering
\caption{Average 3D Euclidean distance test error in the single category experiment for the proposed network with different number of architectures. The best performing architecture is selected for further experimentation. (Scaling is done by 0.01)}
\label{tab1_5}       
\begin{tabular}{p{2.75in}p{0.7in}p{0.7in}}
\hline
\multicolumn{1}{|p{2.75in}}{\multirow{1}{*}{\begin{tabular}{p{2.75in}}\Centering \textbf{Technique}\\\end{tabular}}} & 
\multicolumn{2}{|p{1.4in}|}{\Centering \textbf{3D Euclidean distance}} \\
\hhline{~--}
\multicolumn{1}{|p{2.75in}}{} & 
\multicolumn{1}{|p{0.7in}}{\Centering pred.  \(  \rightarrow  \)  GT} & 
\multicolumn{1}{|p{0.7in}|}{\Centering GT  \(  \rightarrow  \)  pred.} \\
\hhline{---}
\multicolumn{1}{|p{2.75in}}{Lin et al. \cite{original}} & 
\multicolumn{1}{|p{0.7in}}{\Centering 1.768} & 
\multicolumn{1}{|p{0.7in}|}{\Centering 1.763} \\
\hhline{---}
\multicolumn{1}{|p{2.75in}}{Single-encoder 2-decoder network (proposed)} & 
\multicolumn{1}{|p{0.7in}}{\Centering 2.134} & 
\multicolumn{1}{|p{0.7in}|}{\Centering 1.725} \\
\hhline{---}
\multicolumn{1}{|p{2.75in}}{Single-encoder 4-decoder network (proposed)} & 
\multicolumn{1}{|p{0.7in}}{\Centering 1.984} & 
\multicolumn{1}{|p{0.7in}|}{\Centering 1.647} \\
\hhline{---}
\multicolumn{1}{|p{2.75in}}{Single-encoder 8-decoder network (proposed)} & 
\multicolumn{1}{|p{0.7in}}{\Centering \textbf{1.576}} & 
\multicolumn{1}{|p{0.7in}|}{\Centering \textbf{1.553}} \\
\hhline{---}

\end{tabular}
 \end{table*}

\begin{figure*}
\centering
\includegraphics[width=15 cm]{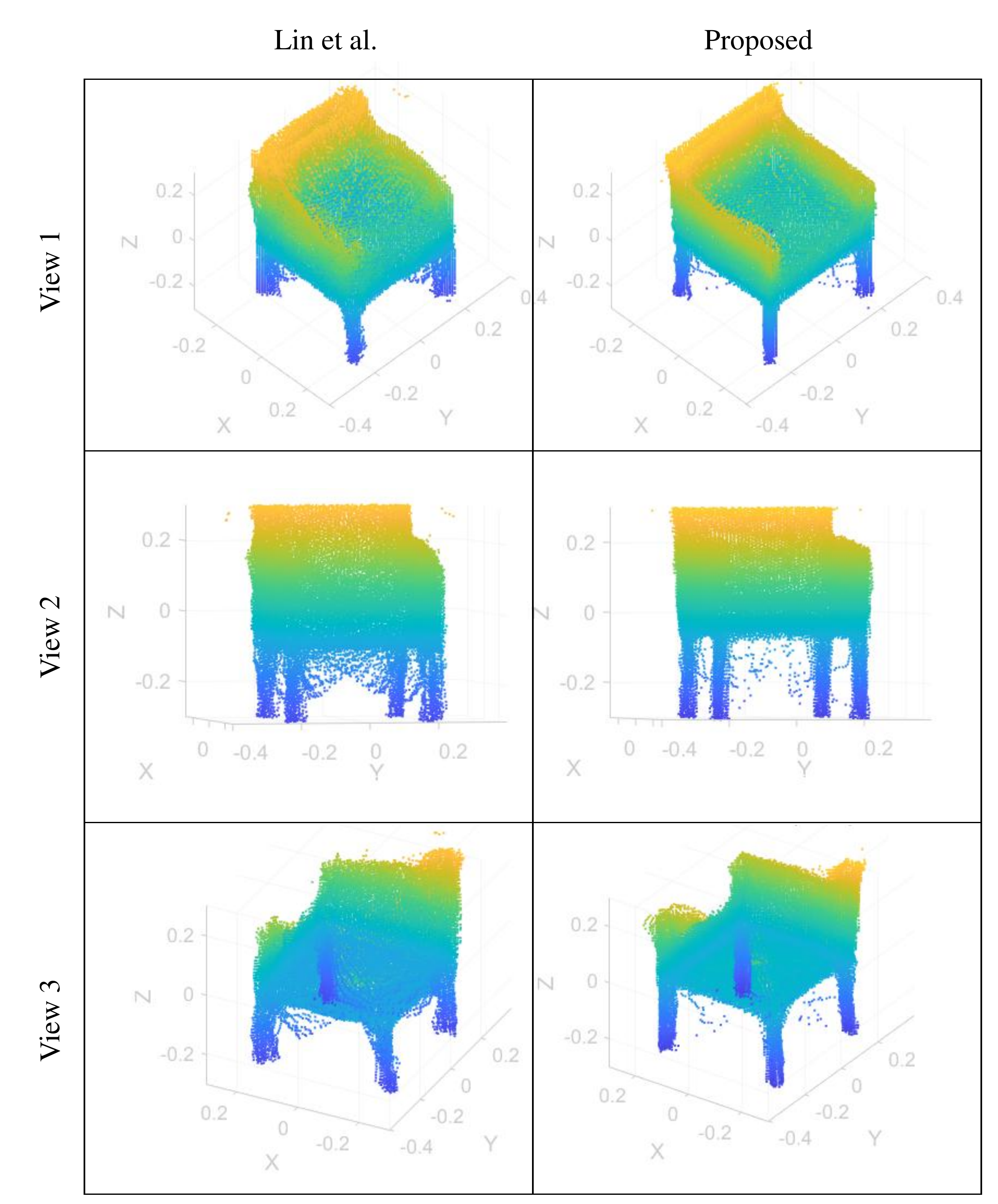}
\caption{Point clouds for a single model in chair category of ShapeNet database as generated by Lin et al. \cite{original} and the proposed technique. The different visualization views show the better point cloud quality of our technique (Best viewed in color).}
\label{fig3a}
\end{figure*}
\begin{figure*}
	\centering
\includegraphics[width=14 cm]{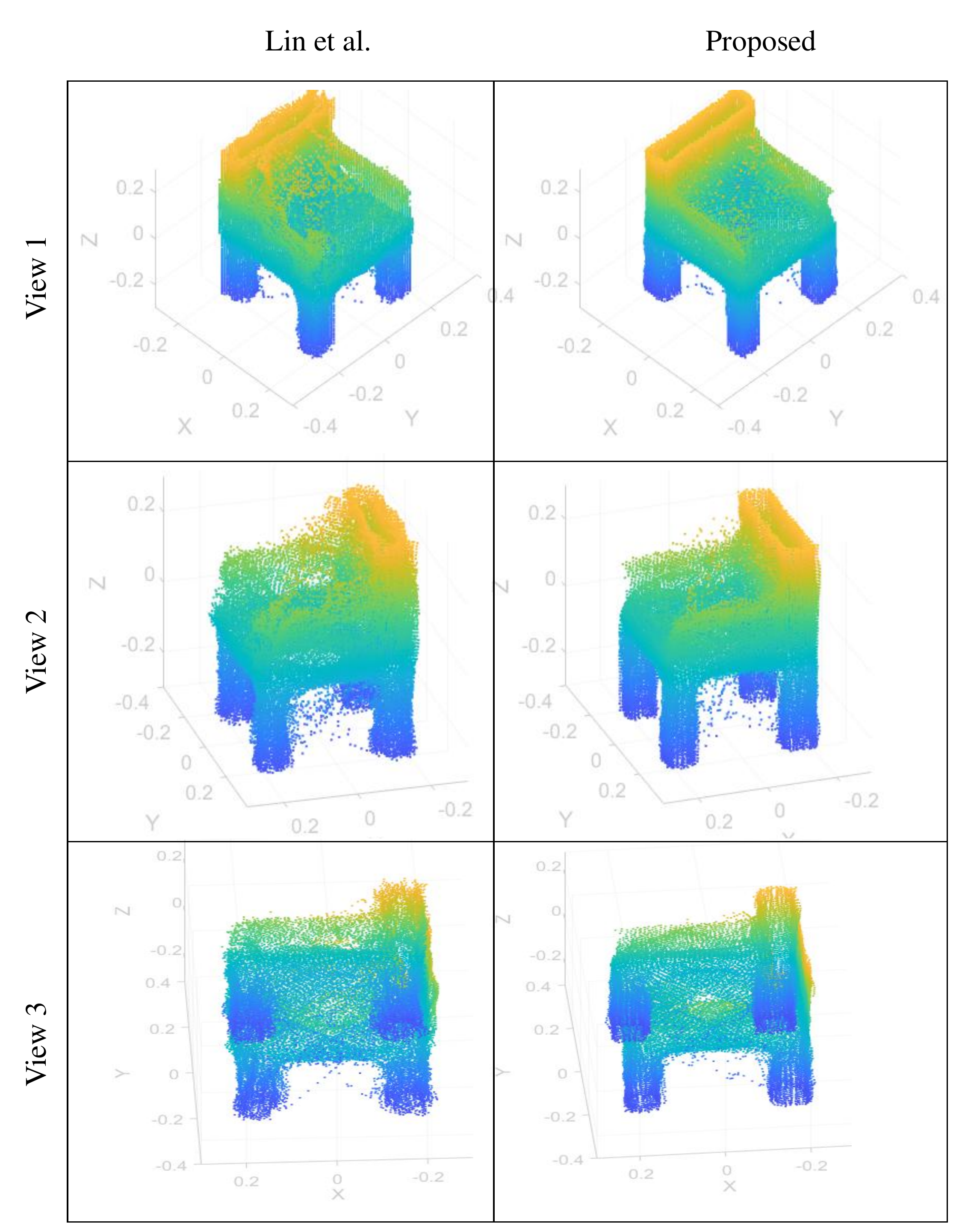}
\caption{Point clouds for another model in chair category of ShapeNet database as generated by Lin et al. \cite{original} and the proposed technique. The different visualization views show the better point cloud quality of the proposed approach (Best viewed in color).}
\label{fig3b}
\end{figure*}

The best performing proposed network variant (i.e., single-encoder 8-decoder proposed model) is used for further experimentation. Next, performance comparison is done against Tatarchenko et al. \cite{o24}, which uses mixed emdedding 3D representation, and Perspective Transformer Networks (PTN) proposed by Yan et al. \cite{t29}, which uses projection error minimization for prediction of volumetric data. The chosen proposed network is pretrained for 200K iterations and fine-tuned end-to-end for 100K iterations. The quantitative results on the testing data are shown in Table \ref{tab2} and in Table \ref{tab3} using different metrics. It is observed from the results in these tables that the proposed technique outperforms all baselines in the metrics used. The results demonstrate that the proposed technique is capable of predicting more accurate 3D representations with higher density and better granularity. The results also show the efficacy of the proposed technique using 2D convolution based model as compared to other 3D convolution based models like PTN \cite{o24}. 2D convolution based architecture has been proved to be more efficient for geometric prediction using a suitable number of viewpoints and combination of the same using geometrical transform functions. Visualization is done for point clouds generated by  the proposed model. The same is shown in Figure \ref{fig3a} and Figure \ref{fig3b}. Compared to Lin et al. \cite{original}, the proposed technique predicts more accurate point clouds with comparable density.


\begin{table*}
	\centering
\caption{Average 3D Euclidean distance test error in the single category experiment. Our technique outperforms all other techniques, which indicates better fine-grained 3D similarity and better point cloud representation. (Scaling is done by 0.01)}
\label{tab2}       
\begin{tabular}{p{2.19in}p{0.91in}p{0.78in}}
\hline
\multicolumn{1}{|p{2.19in}}{\multirow{1}{*}{\begin{tabular}{p{2.19in}}\Centering \textbf{Technique}\\\end{tabular}}} & 
\multicolumn{2}{|p{1.69in}|}{\Centering \textbf{3D Euclidean distance}} \\
\hhline{~--}
\multicolumn{1}{|p{2.19in}}{} & 
\multicolumn{1}{|p{0.91in}}{\Centering pred.  \(  \rightarrow  \)  GT} & 
\multicolumn{1}{|p{0.78in}|}{\Centering GT  \(  \rightarrow  \)  pred.} \\
\hhline{---}
\multicolumn{1}{|p{2.19in}}{3D ConvNet (vol. only) \cite{t29} } & 
\multicolumn{1}{|p{0.91in}}{\Centering 1.827} & 
\multicolumn{1}{|p{0.78in}|}{\Centering 2.660} \\
\hhline{---}
\multicolumn{1}{|p{2.19in}}{PTN (proj. only) \cite{t29} } & 
\multicolumn{1}{|p{0.91in}}{\Centering 2.181} & 
\multicolumn{1}{|p{0.78in}|}{\Centering 2.170} \\
\hhline{---}
\multicolumn{1}{|p{2.19in}}{PTN (vol. $\&$  proj.) \cite{t29} } & 
\multicolumn{1}{|p{0.91in}}{\Centering 1.840} & 
\multicolumn{1}{|p{0.78in}|}{\Centering 2.585} \\
\hhline{---}
\multicolumn{1}{|p{2.19in}}{Tatarchenko et al. \cite{o24} } & 
\multicolumn{1}{|p{0.91in}}{\Centering 2.381} & 
\multicolumn{1}{|p{0.78in}|}{\Centering 3.019} \\
\hhline{---}
\multicolumn{1}{|p{2.19in}}{Lin et al. \cite{original} } & 
\multicolumn{1}{|p{0.91in}}{\Centering 1.768} & 
\multicolumn{1}{|p{0.78in}|}{\Centering 1.763} \\
\hhline{---}
\multicolumn{1}{|p{2.19in}}{Proposed technique} & 
\multicolumn{1}{|p{0.91in}}{\Centering \textbf{1.576}} & 
\multicolumn{1}{|p{0.78in}|}{\Centering \textbf{1.553}} \\
\hhline{---}

\end{tabular}
 \end{table*}



\begin{table*}
	\centering
\caption{Average CD and EMD test errors in single category experiment. Our technique outperforms the best technique, again indicating better fine-grained similarity and better point cloud representation.}
\label{tab3}
\begin{tabular}{p{1.25in}p{1.25in}p{1.25in}}
\hline
\multicolumn{1}{|p{1.25in}}{\Centering \textbf{Technique}} & 
\multicolumn{1}{|p{1.25in}}{\Centering \textbf{Chamfers Distance (CD)}} & 
\multicolumn{1}{|p{1.25in}|}{\Centering \textbf{Earth Mover Distance (EMD)}} \\
\hhline{---}
\multicolumn{1}{|p{1.25in}}{Lin et al. \cite{original} } & 
\multicolumn{1}{|p{1.25in}}{\Centering 6.23} & 
\multicolumn{1}{|p{1.25in}|}{\Centering 9.12} \\
\hhline{---}
\multicolumn{1}{|p{1.25in}}{Proposed technique} & 
\multicolumn{1}{|p{1.25in}}{\Centering \textbf{6.11}} & 
\multicolumn{1}{|p{1.25in}|}{\Centering \textbf{6.38}} \\
\hhline{---}
\end{tabular}
 \end{table*}

\subsubsection{General object categories }

Evaluation is also done via single RGB image to 3D model conversion task using training on multiple ShapeNet categories. Our technique is compared against 3D-R2N2 \cite{o2} using recurrent network based volumetric prediction, Fan et al. \cite{t7} , and Lin et al. \cite{original}. The results of this comparison are reported in Table \ref{tab4}. As it is clear, the proposed technique outperforms all other techniques by impressive margins with better predictions in most cases. Also, the CD and EMD metrics on the multiple-category experiments are listed in Table \ref{tab5}. The proposed technique outperforms others demonstrating its efficacy.

\begin{table*}
\centering
\caption{Average 3D Euclidean distance test error on the multi-category experiment for single-view case (error is shown as: pred  \(  \rightarrow  \)  GT / GT  \(  \rightarrow  \)  pred.). Mean is calculated for all categories. For the single-view generation, the proposed technique outperforms all other techniques (as reported in literature) in 10 and 12 out of 13 categories for the two 3D Euclidean distance metrics. (Scaling is done by 0.01)}
\label{tab4}
\begin{tabular}{p{0.625in}p{0.75in}p{0.75in}p{0.75in}p{0.75in}}
\hline
\multicolumn{1}{|p{0.625in}}{\Centering \textbf{Category}} & 
\multicolumn{1}{|p{0.75in}}{\Centering 3D-R2N2 \cite{o2} } & 
\multicolumn{1}{|p{0.75in}}{\Centering Fan et al.  \cite{t7} } & 
\multicolumn{1}{|p{0.75in}}{\Centering Lin et al. \cite{original} } & 
\multicolumn{1}{|p{0.75in}|}{\Centering Our Technique} \\
\hhline{-----}
\multicolumn{1}{|p{0.625in}}{Airplane} & 
\multicolumn{1}{|p{0.75in}}{\Centering 3.207/2.879} & 
\multicolumn{1}{|p{0.75in}}{\Centering 1.301/1.448} & 
\multicolumn{1}{|p{0.75in}}{\Centering 1.294/1.541} & 
\multicolumn{1}{|p{0.75in}|}{\Centering \textbf{1.103}/\textbf{1.215}} \\
\hhline{-----}
\multicolumn{1}{|p{0.625in}}{Bench} & 
\multicolumn{1}{|p{0.75in}}{\Centering 3.350/3.697} & 
\multicolumn{1}{|p{0.75in}}{\Centering 1.814/1.983} & 
\multicolumn{1}{|p{0.75in}}{\Centering 1.757/1.487} & 
\multicolumn{1}{|p{0.75in}|}{\Centering \textbf{1.514}/\textbf{1.283}} \\
\hhline{-----}
\multicolumn{1}{|p{0.625in}}{Cabinet} & 
\multicolumn{1}{|p{0.75in}}{\Centering 1.636/2.817} & 
\multicolumn{1}{|p{0.75in}}{\Centering 2.463/2.444} & 
\multicolumn{1}{|p{0.75in}}{\Centering 1.814/1.072} & 
\multicolumn{1}{|p{0.75in}|}{\Centering \textbf{1.625}/\textbf{0.976}} \\
\hhline{-----}
\multicolumn{1}{|p{0.625in}}{Car} & 
\multicolumn{1}{|p{0.75in}}{\Centering 1.808/3.238} & 
\multicolumn{1}{|p{0.75in}}{\Centering 1.800/2.053} & 
\multicolumn{1}{|p{0.75in}}{\Centering 1.446/1.061} & 
\multicolumn{1}{|p{0.75in}|}{\Centering \textbf{1.290}/\textbf{0.972}} \\
\hhline{-----}
\multicolumn{1}{|p{0.625in}}{Chair} & 
\multicolumn{1}{|p{0.75in}}{\Centering 2.759/4.207} & 
\multicolumn{1}{|p{0.75in}}{\Centering 1.887/2.355} & 
\multicolumn{1}{|p{0.75in}}{\Centering 1.886/2.041} & 
\multicolumn{1}{|p{0.75in}|}{\Centering \textbf{1.784}/\textbf{1.986}} \\
\hhline{-----}
\multicolumn{1}{|p{0.625in}}{Display} & 
\multicolumn{1}{|p{0.75in}}{\Centering 3.235/4.283} & 
\multicolumn{1}{|p{0.75in}}{\Centering \textbf{1.919}/2.334} & 
\multicolumn{1}{|p{0.75in}}{\Centering 2.142/1.440} & 
\multicolumn{1}{|p{0.75in}|}{\Centering 1.965/\textbf{1.354}} \\
\hhline{-----}
\multicolumn{1}{|p{0.625in}}{Lamp} & 
\multicolumn{1}{|p{0.75in}}{\Centering 8.400/9.722} & 
\multicolumn{1}{|p{0.75in}}{\Centering \textbf{2.347}/\textbf{2.212}} & 
\multicolumn{1}{|p{0.75in}}{\Centering 2.635/4.459} & 
\multicolumn{1}{|p{0.75in}|}{\Centering 2.378/4.101} \\
\hhline{-----}
\multicolumn{1}{|p{0.625in}}{Loudspeaker} & 
\multicolumn{1}{|p{0.75in}}{\Centering 2.652/4.335} & 
\multicolumn{1}{|p{0.75in}}{\Centering 3.215/2.788} & 
\multicolumn{1}{|p{0.75in}}{\Centering 2.371/1.706} & 
\multicolumn{1}{|p{0.75in}|}{\Centering \textbf{2.289}/\textbf{1.655}} \\
\hhline{-----}
\multicolumn{1}{|p{0.625in}}{Rifle} & 
\multicolumn{1}{|p{0.75in}}{\Centering 4.798/2.996} & 
\multicolumn{1}{|p{0.75in}}{\Centering 1.316/1.358} & 
\multicolumn{1}{|p{0.75in}}{\Centering 1.289/1.510} & 
\multicolumn{1}{|p{0.75in}|}{\Centering \textbf{1.201}/\textbf{1.491}} \\
\hhline{-----}
\multicolumn{1}{|p{0.625in}}{Sofa} & 
\multicolumn{1}{|p{0.75in}}{\Centering 2.725/3.628} & 
\multicolumn{1}{|p{0.75in}}{\Centering 2.592/2.784} & 
\multicolumn{1}{|p{0.75in}}{\Centering 1.917/1.423} & 
\multicolumn{1}{|p{0.75in}|}{\Centering \textbf{1.752}/\textbf{1.324}} \\
\hhline{-----}
\multicolumn{1}{|p{0.625in}}{Table} & 
\multicolumn{1}{|p{0.75in}}{\Centering 3.118/4.208} & 
\multicolumn{1}{|p{0.75in}}{\Centering 1.874/2.229} & 
\multicolumn{1}{|p{0.75in}}{\Centering 1.689/1.620} & 
\multicolumn{1}{|p{0.75in}|}{\Centering \textbf{1.435}/\textbf{1.412}} \\
\hhline{-----}
\multicolumn{1}{|p{0.625in}}{Telephone} & 
\multicolumn{1}{|p{0.75in}}{\Centering 2.202/3.314} & 
\multicolumn{1}{|p{0.75in}}{\Centering \textbf{1.516}/1.989} & 
\multicolumn{1}{|p{0.75in}}{\Centering 1.939/1.198} & 
\multicolumn{1}{|p{0.75in}|}{\Centering 1.754/\textbf{1.154}} \\
\hhline{-----}
\multicolumn{1}{|p{0.625in}}{Watercraft} & 
\multicolumn{1}{|p{0.75in}}{\Centering 3.592/4.007} & 
\multicolumn{1}{|p{0.75in}}{\Centering 1.715/1.877} & 
\multicolumn{1}{|p{0.75in}}{\Centering 1.813/1.550} & 
\multicolumn{1}{|p{0.75in}|}{\Centering \textbf{1.686}/\textbf{1.435}} \\
\hhline{-----}
\multicolumn{1}{|p{0.625in}}{\textbf{Mean}} & 
\multicolumn{1}{|p{0.75in}}{\Centering 3.345/4.102} & 
\multicolumn{1}{|p{0.75in}}{\Centering 1.982/2.146} & 
\multicolumn{1}{|p{0.75in}}{\Centering 1.846/1.701} & 
\multicolumn{1}{|p{0.75in}|}{\Centering \textbf{1.675/1.568}} \\
\hhline{-----}

\end{tabular}
\end{table*}



\begin{table*}
	\centering
\caption{Single-view pixel cloud generation results on multi-category experiment. The metrics are computed on 1024 points after alignment of the generated point clouds with their respective GT point clouds. Scaling is done by a factor of 100. Mean is calculated for all categories. For the single-view generation, our technique outperforms others (as reported in literature) in 9 out of 13 categories (for CD metric), and 10 out of 13 categories (for EMD metric).}
\label{tab5}       
\begin{tabular}{p{0.7in}p{0.45in}p{0.45in}p{0.45in}p{0.45in}p{0.45in}p{0.45in}}
\hline
\multicolumn{1}{|p{0.7in}}{\multirow{1}{*}{\begin{tabular}{p{0.7in}}\Centering \textbf{Category}\\\end{tabular}}} & 
\multicolumn{3}{|p{1.35in}}{\Centering \textbf{CD}} & 
\multicolumn{3}{|p{1.35in}|}{\Centering \textbf{EMD}} \\
\hhline{~------}
\multicolumn{1}{|p{0.7in}}{} & 
\multicolumn{1}{|p{0.7in}}{\Centering Lin et al. \cite{original} } & 
\multicolumn{1}{|p{0.8in}}{\Centering 3D-LMNet \cite{lmnet} } & 
\multicolumn{1}{|p{0.7in}}{\Centering Our Technique} & 
\multicolumn{1}{|p{0.7in}}{\Centering Lin et al. \cite{original} } & 
\multicolumn{1}{|p{0.8in}}{\Centering 3D-LMNet \cite{lmnet} } & 
\multicolumn{1}{|p{0.7in}|}{\Centering Our Technique} \\
\hhline{-------}
\multicolumn{1}{|p{0.7in}}{Airplane} & 
\multicolumn{1}{|p{0.45in}}{\Centering 3.68} & 
\multicolumn{1}{|p{0.45in}}{\Centering 3.34} & 
\multicolumn{1}{|p{0.45in}}{\Centering \textbf{3.11}} & 
\multicolumn{1}{|p{0.45in}}{\Centering 5.64} & 
\multicolumn{1}{|p{0.45in}}{\Centering \textbf{4.77}} & 
\multicolumn{1}{|p{0.45in}|}{\Centering 4.78} \\
\hhline{-------}
\multicolumn{1}{|p{0.7in}}{Bench} & 
\multicolumn{1}{|p{0.45in}}{\Centering 4.57} & 
\multicolumn{1}{|p{0.45in}}{\Centering 4.55} & 
\multicolumn{1}{|p{0.45in}}{\Centering \textbf{4.34}} & 
\multicolumn{1}{|p{0.45in}}{\Centering 5.76} & 
\multicolumn{1}{|p{0.45in}}{\Centering 4.99} & 
\multicolumn{1}{|p{0.45in}|}{\Centering \textbf{4.61}} \\
\hhline{-------}
\multicolumn{1}{|p{0.7in}}{Cabinet} & 
\multicolumn{1}{|p{0.45in}}{\Centering 6.76} & 
\multicolumn{1}{|p{0.45in}}{\Centering 6.09} & 
\multicolumn{1}{|p{0.45in}}{\Centering \textbf{5.89}} & 
\multicolumn{1}{|p{0.45in}}{\Centering \textbf{6.01}} & 
\multicolumn{1}{|p{0.45in}}{\Centering 6.35} & 
\multicolumn{1}{|p{0.45in}|}{\Centering 6.37} \\
\hhline{-------}
\multicolumn{1}{|p{0.7in}}{Car} & 
\multicolumn{1}{|p{0.45in}}{\Centering 4.95} & 
\multicolumn{1}{|p{0.45in}}{\Centering 4.55} & 
\multicolumn{1}{|p{0.45in}}{\Centering \textbf{4.52}} & 
\multicolumn{1}{|p{0.45in}}{\Centering 4.38} & 
\multicolumn{1}{|p{0.45in}}{\Centering \textbf{4.10}} & 
\multicolumn{1}{|p{0.45in}|}{\Centering 4.11} \\
\hhline{-------}
\multicolumn{1}{|p{0.7in}}{Chair} & 
\multicolumn{1}{|p{0.45in}}{\Centering 6.45} & 
\multicolumn{1}{|p{0.45in}}{\Centering \textbf{6.41}} & 
\multicolumn{1}{|p{0.45in}}{\Centering 6.47} & 
\multicolumn{1}{|p{0.45in}}{\Centering 9.25} & 
\multicolumn{1}{|p{0.45in}}{\Centering 8.02} & 
\multicolumn{1}{|p{0.45in}|}{\Centering \textbf{6.53}} \\
\hhline{-------}
\multicolumn{1}{|p{0.7in}}{Display} & 
\multicolumn{1}{|p{0.45in}}{\Centering \textbf{6.27}} & 
\multicolumn{1}{|p{0.45in}}{\Centering 6.40} & 
\multicolumn{1}{|p{0.45in}}{\Centering 6.36} & 
\multicolumn{1}{|p{0.45in}}{\Centering 7.47} & 
\multicolumn{1}{|p{0.45in}}{\Centering 7.13} & 
\multicolumn{1}{|p{0.45in}|}{\Centering \textbf{6.74}} \\
\hhline{-------}
\multicolumn{1}{|p{0.7in}}{Lamp} & 
\multicolumn{1}{|p{0.45in}}{\Centering \textbf{6.25}} & 
\multicolumn{1}{|p{0.45in}}{\Centering 7.10} & 
\multicolumn{1}{|p{0.45in}}{\Centering 7.08} & 
\multicolumn{1}{|p{0.45in}}{\Centering 16.12} & 
\multicolumn{1}{|p{0.45in}}{\Centering 15.80} & 
\multicolumn{1}{|p{0.45in}|}{\Centering \textbf{12.11}} \\
\hhline{-------}
\multicolumn{1}{|p{0.7in}}{Loudspeaker} & 
\multicolumn{1}{|p{0.45in}}{\Centering 8.72} & 
\multicolumn{1}{|p{0.45in}}{\Centering 8.10} & 
\multicolumn{1}{|p{0.45in}}{\Centering \textbf{7.92}} & 
\multicolumn{1}{|p{0.45in}}{\Centering 8.92} & 
\multicolumn{1}{|p{0.45in}}{\Centering 9.15} & 
\multicolumn{1}{|p{0.45in}|}{\Centering \textbf{7.86}} \\
\hhline{-------}
\multicolumn{1}{|p{0.7in}}{Rifle} & 
\multicolumn{1}{|p{0.45in}}{\Centering 2.87} & 
\multicolumn{1}{|p{0.45in}}{\Centering \textbf{2.75}} & 
\multicolumn{1}{|p{0.45in}}{\Centering 2.81} & 
\multicolumn{1}{|p{0.45in}}{\Centering 8.21} & 
\multicolumn{1}{|p{0.45in}}{\Centering 6.08} & 
\multicolumn{1}{|p{0.45in}|}{\Centering \textbf{5.89}} \\
\hhline{-------}
\multicolumn{1}{|p{0.7in}}{Sofa} & 
\multicolumn{1}{|p{0.45in}}{\Centering 6.34} & 
\multicolumn{1}{|p{0.45in}}{\Centering 5.85} & 
\multicolumn{1}{|p{0.45in}}{\Centering \textbf{5.69}} & 
\multicolumn{1}{|p{0.45in}}{\Centering 6.77} & 
\multicolumn{1}{|p{0.45in}}{\Centering 5.65} & 
\multicolumn{1}{|p{0.45in}|}{\Centering \textbf{5.21}} \\
\hhline{-------}
\multicolumn{1}{|p{0.7in}}{Table} & 
\multicolumn{1}{|p{0.45in}}{\Centering 6.12} & 
\multicolumn{1}{|p{0.45in}}{\Centering 6.05} & 
\multicolumn{1}{|p{0.45in}}{\Centering \textbf{5.62}} & 
\multicolumn{1}{|p{0.45in}}{\Centering 8.32} & 
\multicolumn{1}{|p{0.45in}}{\Centering 7.82} & 
\multicolumn{1}{|p{0.45in}|}{\Centering \textbf{6.14}} \\
\hhline{-------}
\multicolumn{1}{|p{0.7in}}{Telephone} & 
\multicolumn{1}{|p{0.45in}}{\Centering 4.72} & 
\multicolumn{1}{|p{0.45in}}{\Centering 4.63} & 
\multicolumn{1}{|p{0.45in}}{\Centering \textbf{4.51}} & 
\multicolumn{1}{|p{0.45in}}{\Centering 6.23} & 
\multicolumn{1}{|p{0.45in}}{\Centering 5.43} & 
\multicolumn{1}{|p{0.45in}|}{\Centering \textbf{5.11}} \\
\hhline{-------}
\multicolumn{1}{|p{0.7in}}{Watercraft} & 
\multicolumn{1}{|p{0.45in}}{\Centering 4.42} & 
\multicolumn{1}{|p{0.45in}}{\Centering 4.37} & 
\multicolumn{1}{|p{0.45in}}{\Centering \textbf{4.24}} & 
\multicolumn{1}{|p{0.45in}}{\Centering 6.14} & 
\multicolumn{1}{|p{0.45in}}{\Centering 5.68} & 
\multicolumn{1}{|p{0.45in}|}{\Centering \textbf{5.25}} \\
\hhline{-------}
\multicolumn{1}{|p{0.7in}}{\textbf{Mean}} & 
\multicolumn{1}{|p{0.45in}}{\Centering 5.55} & 
\multicolumn{1}{|p{0.45in}}{\Centering 5.40} & 
\multicolumn{1}{|p{0.45in}}{\Centering \textbf{4.70}} & 
\multicolumn{1}{|p{0.45in}}{\Centering 7.64} & 
\multicolumn{1}{|p{0.45in}}{\Centering 7.00} & 
\multicolumn{1}{|p{0.45in}|}{\Centering \textbf{6.21}} \\
\hhline{-------}
\end{tabular}
\end{table*}


\section{Conclusion}
\label{Con}
In this work, a new deep network architecture was proposed for generation of 3D point clouds which are powerful 3D model representation modelss. The 2D convolution based deep network aims to introduce parallelization by using a single-encoder multiple-decoder structure. The proposed network is tested on the popular ShapeNet database used for benchmarking of state of the art 2D RGB image to 3D model conversion techniques. The database has been used by many researchers. Compared to state of the art techniques, the proposed technique is experimentally proved to be more efficient and robust. The same is illustrated in the form of various tabular experimental results and graphical point cloud comparisons. Our proposed network can be used in other computer vision tasks as it is simple, robust and efficient due to its parallelized architecture as found in expert image processing systems like those of humans. In the future, the parallelization concept will be investigated in other strategies for 3D model generation with more complex datasets.

\textbf{Conflict of interest}
The authors declare no conflict of interest.

\bibliographystyle{spmpsci}      
\bibliography{Manuscript}   

\end{document}